\documentclass[a4paper,graybox]{svmult}
\usepackage{makeidx}         % allows index generation
\usepackage{graphicx}
\usepackage{multicol}
\usepackage[bottom]{footmisc}
\usepackage[tight]{subfigure}
\usepackage{newtxtext}
\usepackage{newtxmath} % selects Times Roman as basic font
\usepackage{lipsum}
\usepackage{enumitem}
\usepackage{amsmath}
\usepackage{mathtools}
\usepackage{pifont}
\usepackage{indentfirst}
\usepackage{hyperref}
\usepackage{bm}
\usepackage{cite}
\usepackage[a4paper,top=91pt,left=133pt,right=133pt,bottom=176pt]{geometry}

\makeindex

\newlist{todolist}{itemize}{2}
\setlist[todolist]{label=$\square$}

\begin{document}
% \title*{Exploration on Bayesian Hilbert Map with Pareto Monte Carlo Tree Search}
\title*{Multi-Objective Autonomous Exploration on Real-Time Continuous Occupancy Maps}
\author{Zheng Chen$^1$, Weizhe Chen$^1$, Shi Bai$^2$, Lantao Liu$^1$}
\institute{$^1$ Z. Chen, W. Chen and L. Liu are with the Luddy School of Informatics, Computing and Engineering, Indiana University-Bloomington. \\
$^2$ S. Bai is with Wing, Alphabet Inc. 
}
\maketitle
\thispagestyle{empty}
\vspace{-8em}
% \abstract{
% %=== Autonomous exploration is important ===
% Autonomous exploration in unknown environments using mobile robots is the pillar of many robotic applications.
% %=== What problem are we solving? ===
% Existing exploration frameworks either select the nearest geometric frontier or the nearest information-theoretic frontier.
% However, \textit{just because a frontier itself is informative does not necessarily mean that the robot will be in an informative area after reaching that frontier}.
% %=== Our solution ===
% To fill this gap, we propose to use a multi-objective variant of Monte-Carlo tree search that provides a non-myopic Pareto optimal action sequence leading the robot to a frontier with greatest extent of unknown area uncovering.
% We also adopted Bayesian Hilbert Map (BHM) for continuous occupancy mapping and made it more applicable to real-time tasks.
% % TODO: no need for two-stage planning
% %=== Results and take-home messages ===
% \WZ{Simulation and field experiments to validate the effectiveness and efficiency of our method?}
% }
\section{Motivation, Problem Statement, Related Work} \label{sec:introduction}
%=== Pull figure ===
\begin{figure}[htbp] \vspace{-15pt}
    \centering
    % Nearest geometric frontier
    \begin{minipage}[t]{0.3\linewidth}
        \centering
        \includegraphics[width=1\linewidth,height=1\linewidth]{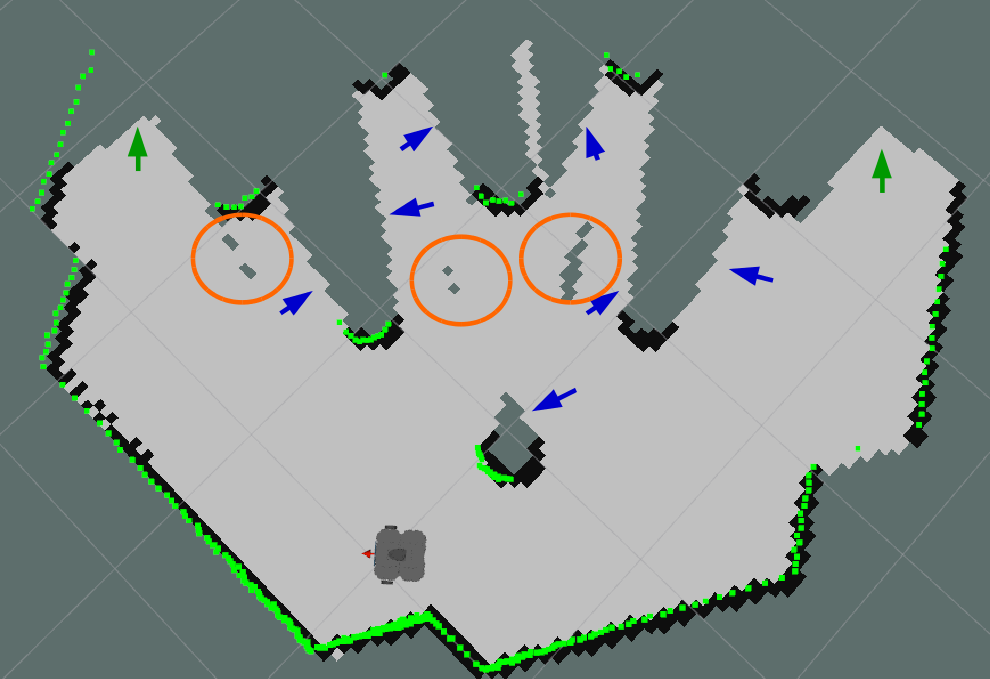}
        \caption{
            Discrete occupancy grid map with frontiers pointed by arrows.
            % Exploring frontiers pointed by green arrows will uncover larger unknown area than that of the blue ones.
            % Inconsistent occupancy predictions are circled by orange, which motivates us to embrace continuous occupancy mapping.
        }
        \label{fig:pull_figure_1}
    \end{minipage}
    % % Most informative frontier
    % \begin{minipage}[t]{0.3\linewidth}
    %     \centering
    %     \includegraphics[width=1\linewidth,height=1\linewidth]{figs/gp_frontiers.png}
    %     \caption{Informative locations}
    %     \label{fig:pull_figure_2}
    % \end{minipage}
    % Informative area
    \quad \quad 
    \begin{minipage}[t]{0.3\linewidth}
        \centering
        \includegraphics[width=1\linewidth,height=1\linewidth]{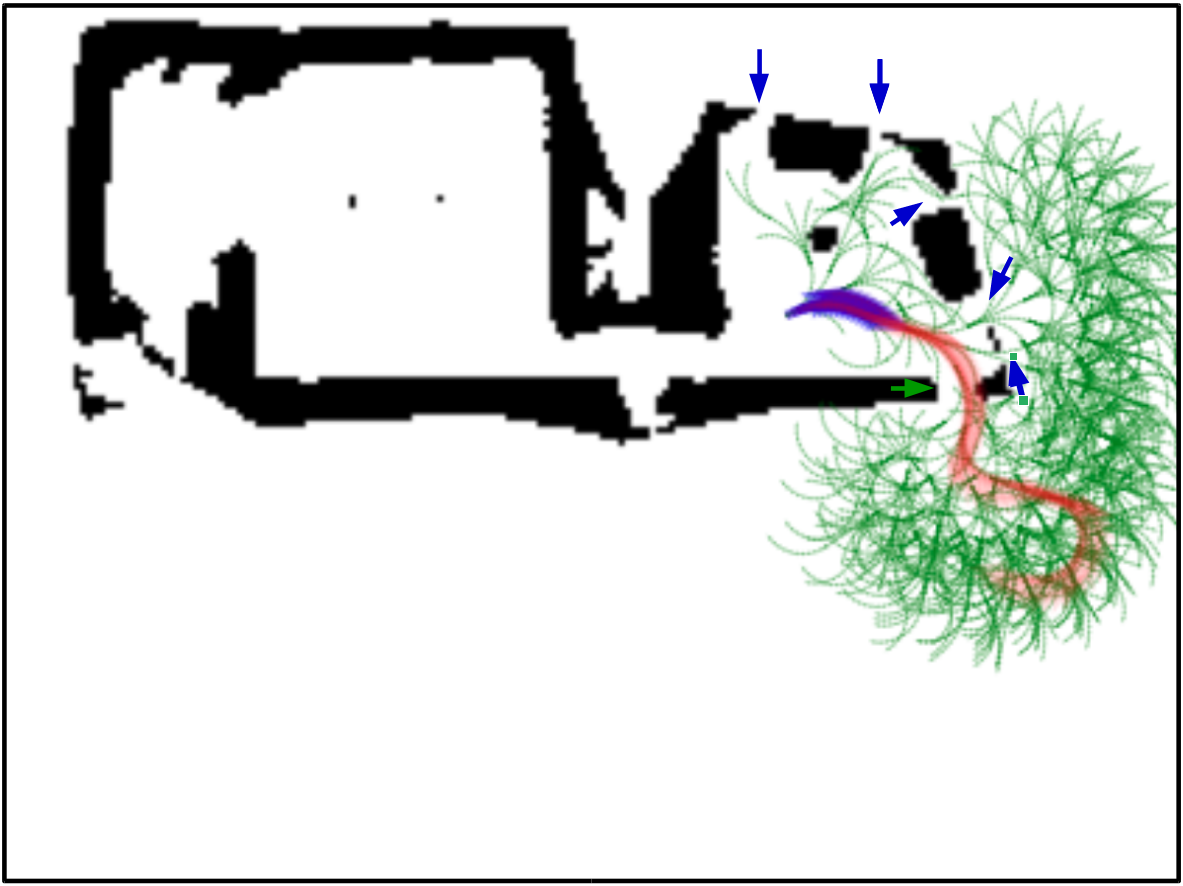}
        \caption{Exploring frontiers that lead to large unknown areas.}
        \label{fig:pull_figure_3}
    \end{minipage} %\vspace{-15pt}
\end{figure}

%=== Autonomous exploration is important ===
Autonomous exploration in unknown environments is a needed capability for many robotic applications.
%=== What problem are we solving? ===
Existing exploration frameworks  rely on either geometric frontiers~\cite{yamauchi1997frontier,burgard2000collaborative,shen2012stochastic,cieslewski2017rapid} or  information-theoretic frontiers~\cite{charrow2015information,zhang2019fsmi,francis_functional_2020,francis_occupancy_2019,bai_information-theoretic_2016,jadidi_mutual_2015}, with which separate planners are required to plan collision-free paths towards the selected frontiers.
However, \textit{just because a frontier itself is informative does not necessarily mean that the robot will be in an informative ``area" after reaching that frontier}.
For example, in Fig.\,\ref{fig:pull_figure_1}, exploring the green arrow pointed open areas will uncover more unknown spaces compared to visiting those indicated by blue arrows.
Thus the frontiers of blue arrows will lead to ``less informative" actions compared with those of the green arrows.
%=== Our solution ===
In this work, we propose to use a multi-objective variant of Monte-Carlo tree search --- ParetoMCTS --- that provides a non-myopic Pareto optimal action sequence and hence leads the robot to frontiers with the greatest extent of unknown areal uncovering (Fig.\,\ref{fig:pull_figure_3} demonstrates that our planner prefers the frontier at the bottom because it leads to a larger uncovered space).
% The red dots in Fig.\,\ref{fig:pull_figure_1} and the bright area in Fig\,\ref{fig:pull_figure_2} are frontiers derived from a discrete occupancy map and a continuous occupancy map, respectively.
%
Also importantly, some ``fake'' frontiers (circled in orange) can present in the discrete occupancy grid map (Fig.\,\ref{fig:pull_figure_1}) due to the underlying inappropriate assumption that grids are independent to each other.
%To avoid these inconsistent results by taking into consideration the inter-dependencies among occupancy values, 
To fix such issue, we need to explicitly model the inter-dependencies among occupancy values, and hereby we adopt the Bayesian Hilbert Map (BHM)~\cite{senanayake2017bayesian} for continuous occupancy mapping which, however, is computationally prohibitive for large complex problems. 
We propose strategies and preliminary results for addressing above issues for real-time environment exploration applications. %further made it more applicable to real-time tasks.
\section{Technical Approach}\label{sec:approach} %\vspace{-10pt}
%=== Bayesian Hilbert Map ===
Bayesian Hilbert Map~\cite{senanayake2017bayesian} is an extension of Hilbert Map~\cite{ramos2016hilbert}, which represents the environment with a \textit{continuous occupancy map} by using a logistic regression classifier in a Hilbert space.
The probability of a query location $\mathbf{x}$ being occupied, i.e., $y=1$, can be computed by
% The occupancy prediction model in BHM is defined using a sigmoid function:
\begin{equation}
    \label{eq:logi}
    p(y=1|\mathbf{x}, \mathbf{w}) = \left ( 1 + \exp(\mathbf{w}^T\Phi(\mathbf{x})) \right )^{-1} = \sigma(-\mathbf{w}^T\Phi(\mathbf{x})),
\end{equation}
where $\mathbf{w}$ is the model parameters to be optimized, $\Phi(\mathbf{x})$ is a radial basis function (RBF) feature transformation of $\mathbf{x}$ centered at some spatially fixed points called hinge points, and $\sigma(\cdot)$ represents the logit function.
% where $y=1$ means the probability in Eq.~\eqref{eq:logi} is an {\em occupancy} probability; $\mathbf{x}$ is the query point, which is represented as the coordinate of the point that we want to know the occupancy; $\mathbf{w}$ is the model parameters which need to be optimized during training; $\Phi(\cdot)$ is a new feature vector which is transformed from the original input vector $\mathbf{x}$.
%
As a Bayesian approach, BHM places a factorized Gaussian prior $p(\mathbf{w}|\mathbf{\alpha})=\mathcal{N}\left(\mathbf{w}|\mathbf{0},\operatorname{diag}(\mathbf{\alpha})\right)$ on the model parameter and infers its posterior.
Analytical posterior is intractable due to the nonconjugacy between the Gaussian prior and Bernoulli likelihood.
% However, since the likelihood function is a sigmoid function, no analytical solution for the posterior distribution $P(\mathbf{w}|\mathbf{x}, \mathbf{y})$ could be obtained.
Therefore, the posterior $p(\mathbf{w}, \mathbf{\alpha}|\mathbf{x}, \mathbf{y})$ is approximated by a variational distribution $q(\mathbf{w}, \mathbf{\alpha})$, and we implicitly minimize the Kullback-Leibler (KL) divergence between $q$ and $p$ so that the approximate distribution is close to the true posterior~\cite{senanayake2017bayesian}.
% \begin{equation}
%     \label{eq:posterior}
%     Q(\mathbf{w}, \alpha) \approx P(\mathbf{w}|\mathbf{x}, \mathbf{y}) = \frac{P(\mathbf{y}|\mathbf{x}, \mathbf{w})P(\mathbf{w|\alpha})P(\alpha)}{P(\mathbf{y})},
% \end{equation}
% where $\alpha$ is an additionally introduced parameter due to the approximation. A variational inference method was used to learn the parameters.
% More details can be found in .

%=== Monte-Carlo tree search part ===
Monte Carlo tree search (MCTS) methods belong to best-first search algorithms which expand the most promising subtrees first.
We have been developing the Pareto Monte Carlo tree search (ParetoMCTS)~\cite{chen2019pareto}, which is a multi-objective extension of MCTS that seeks for the {\em Pareto optimal decisions} where any objective cannot be further improved without hurting (or compromising) other objectives. % --- Pareto optimality.
Starting from the root node, we select the child node according to Pareto Upper Confidence Bound (UCB) criterion at each level until an expandable node with unexpanded children is encountered.
A random child node is expanded and its value is evaluated through forward simulation.
We update the statistics in all the visited nodes using the estimated value to bias the searching process in the next round.
\begin{figure}[htbp] %\vspace{-10pt}
    \centering
    \includegraphics[width=1\linewidth]{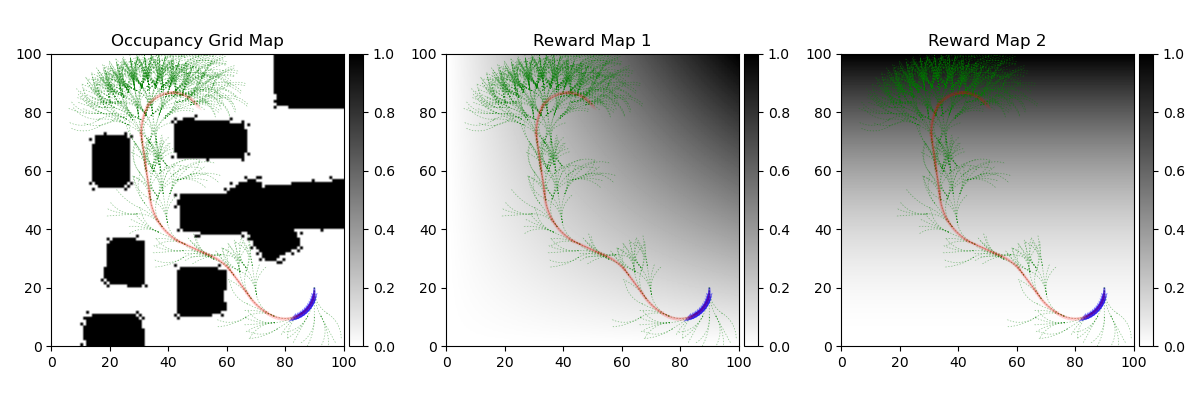}
    \caption{Demonstration of searching an informative trajectory in a confined environment given two exploration reward maps.
    The green dots represent the resulting search tree.
    Red line shows the best trajectory and blue line is the best next action.
    Reward map 1 (middle figure with grey gradient color) indicates that the upper-right corner has higher exploration reward while the reward map 2  (right) suggests that the upper part is worth visiting.
    The tree expansion is first biased towards the high reward area, and then it starts scattering in order not to compromise any objective. \vspace{-10pt}
    }
    \label{fig:pareto_mcts}
\end{figure}
Fig.\,\ref{fig:pareto_mcts} shows the result of a ParetoMCTS when optimizing two exploration objectives in a confined environment.
ParetoMCTS grows an asymmetric tree towards the high-reward area and tries to optimize both objectives.
The resulting tree is collision-free and satisfies robot's motion constraints.
%For more details about ParetoMCTS, we refer the reader to \cite{chen2019pareto}.

In this paper, the inputs of ParetoMCTS include current robot pose, a occupancy map, a entropy map and a frontier dynamics map.
% we assume the localization could be perfectly accessed at any time. Since we can have an occupancy map with a specific resolution from BHM, we can compute a respective entropy map with the specified resolution from:
Given a grid $\mathbf{m}_i$ and its occupancy probability $P_i$, the entropy is given by
\begin{equation}
    \label{eq:entropy}
    H(\mathbf{m}_i) = -P_i\log(P_i) - (1-P_i)\log(1-P_i).
\end{equation}
% where $\mathbf{m}_i$ is the $i^{th}$ query point while $p_i$ is the predicted occupancy probability at $\mathbf{m}_i$.
We defined the frontier dynamics map to be the absolute difference between two consecutive occupancy maps.
% The frontier dynamics map is computed by extracting the difference between occupancy maps at two consecutive time steps.

ParetoMCTS outputs the most informative trajectory (red line in Fig.\,\ref{fig:pareto_mcts}) for exploration given the current perception information.
One can simply follow this trajectory and plan again when reaching the end.
However, considering that the perception information is continuously updating, we only take the first action (blue line in Fig.\,\ref{fig:pareto_mcts}) in the resulting trajectory and replan in a receding horizon manner to avoid newly encountered obstacles and keep the informative trajectory up to date.
\vspace{-15pt}
\section{Results}\label{sec:results}
%\vspace{-10pt}
In this section, we demonstrate that by simultaneously considering two (potentially reward-diverging) objectives -- maximization of frontier dynamics and minimization of map entropy, our system could efficiently explore unknown environments in real time even though our map is continuous. We use BHM to build the occupancy map and reward maps for ParetoMCTS, and use OctoMap~\cite{hornung2013octomap} to build a 3D map. 

\begin{figure}[h]
\centering
    \begin{minipage}[tb]{0.25\linewidth}
        \centering
        \includegraphics[width=1\linewidth]{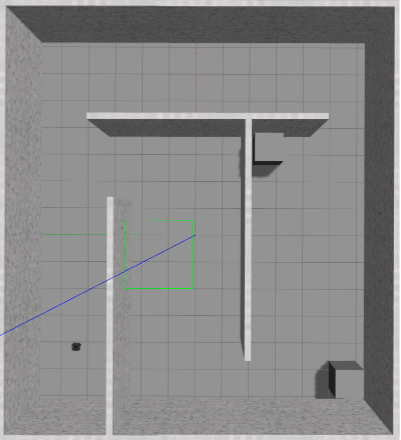}\\(a)
        \label{fig:env}
    \end{minipage}
    \begin{minipage}[tb]{0.25\linewidth}
        \centering
        \includegraphics[width=1\linewidth]{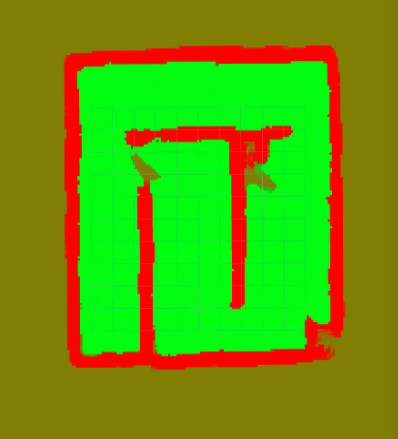}\\(b)
        \label{fig:env_bhm}
    \end{minipage}
    \begin{minipage}[tb]{0.25\linewidth}
        \centering
        \includegraphics[width=1\linewidth]{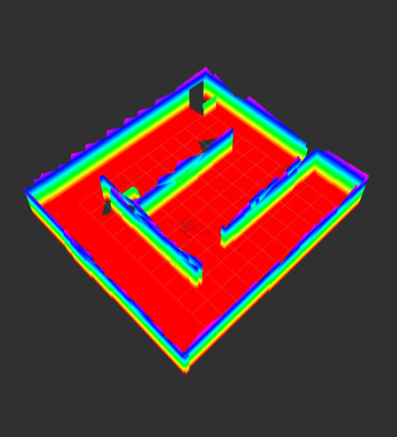}\\(c)
        \label{fig:env_3d}
    \end{minipage}
    \caption{\small
        Exploration results in a simulated environment. (a)~The simulated environment that needs to be explored. (b)~The mapping results from BHM. Green, red, brown colors represent free, occupied and uncertain spaces, respectively. (c)~3D view of the mapping result. %Octomap. %Colors indicate height values.
        \vspace{-10pt}
    }
    \label{fig:exploration_results}
\end{figure}

\begin{figure}[h] \vspace{-10pt}
    \begin{minipage}[tb]{0.5\linewidth}
        \centering
        \includegraphics[width=0.98\linewidth]{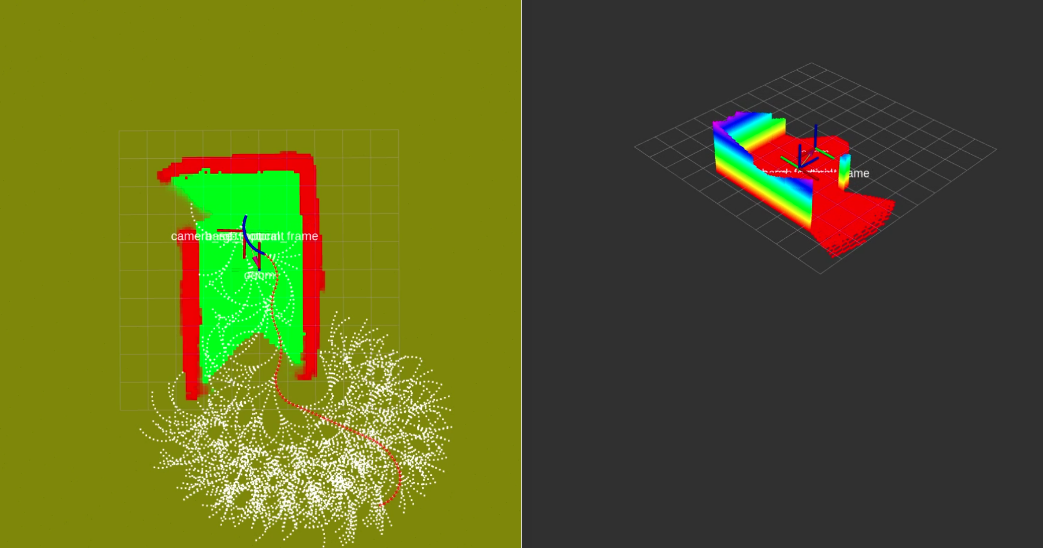}\\(a)
        \label{fig:test_1}
    \end{minipage}
    \begin{minipage}[tb]{0.5\linewidth}
        \centering
        \includegraphics[width=0.98\linewidth]{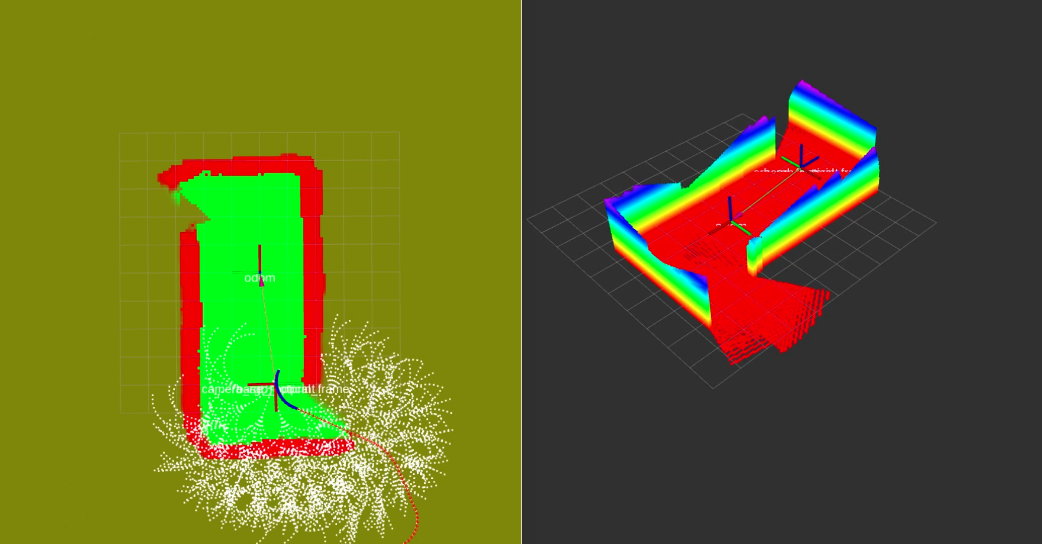}\\(b)
        \label{fig:test_2}
    \end{minipage}
    \begin{minipage}[tb]{0.5\linewidth}
        \centering
        \includegraphics[width=0.98\linewidth]{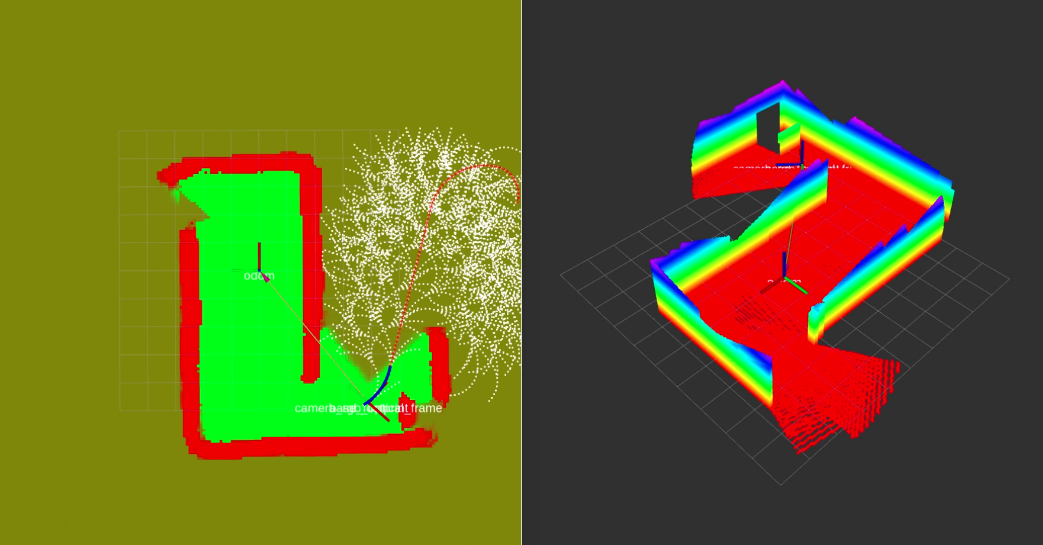}\\(c)
        \label{fig:test_3}
    \end{minipage}
    \begin{minipage}[tb]{0.5\linewidth}
        \centering
        \includegraphics[width=0.98\linewidth]{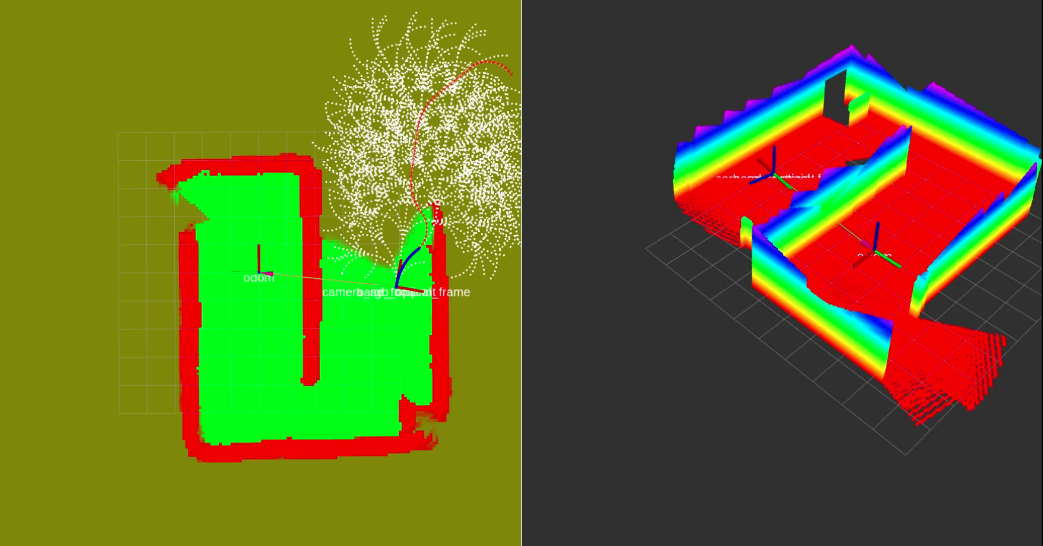}\\(d)
        \label{fig:test_4}
    \end{minipage}
    \begin{minipage}[tb]{0.5\linewidth}
        \centering
        \includegraphics[width=0.98\linewidth]{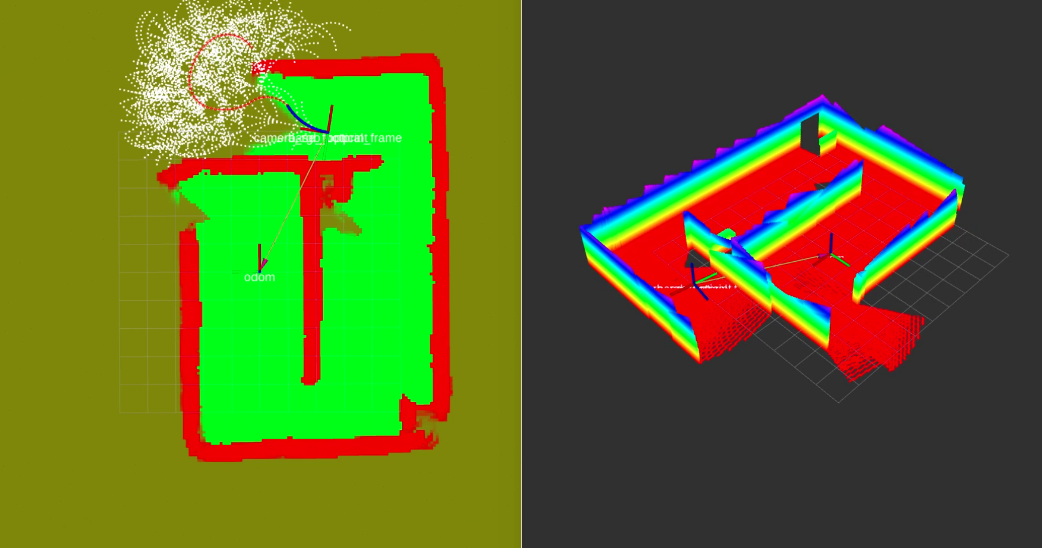}\\(e)
        \label{fig:test_5}
    \end{minipage}
    \begin{minipage}[tb]{0.5\linewidth}
        \centering
        \includegraphics[width=0.98\linewidth]{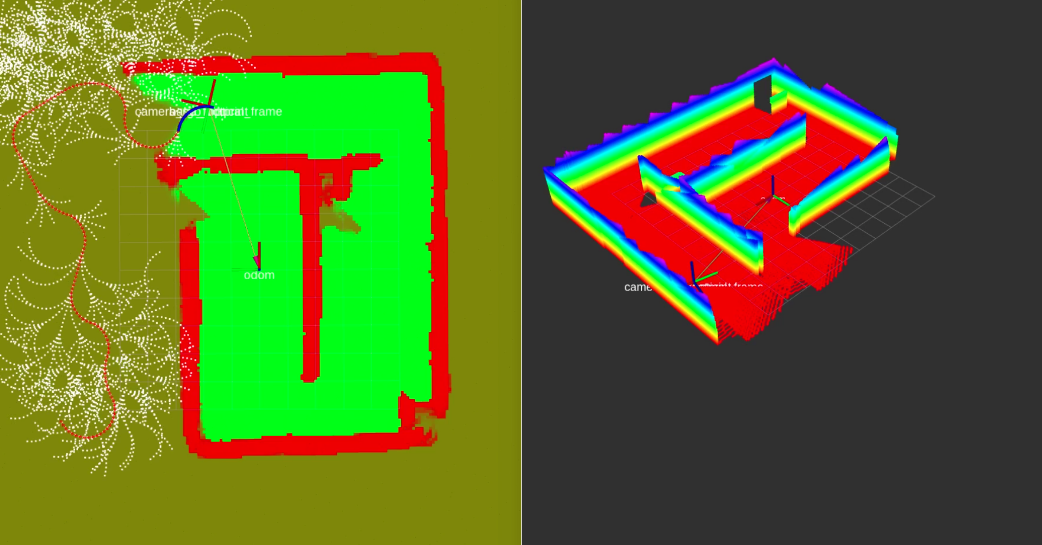}\\(f)
        \label{fig:test_6}
    \end{minipage}
    \begin{minipage}[tb]{0.5\linewidth}
        \centering
        \includegraphics[width=0.98\linewidth]{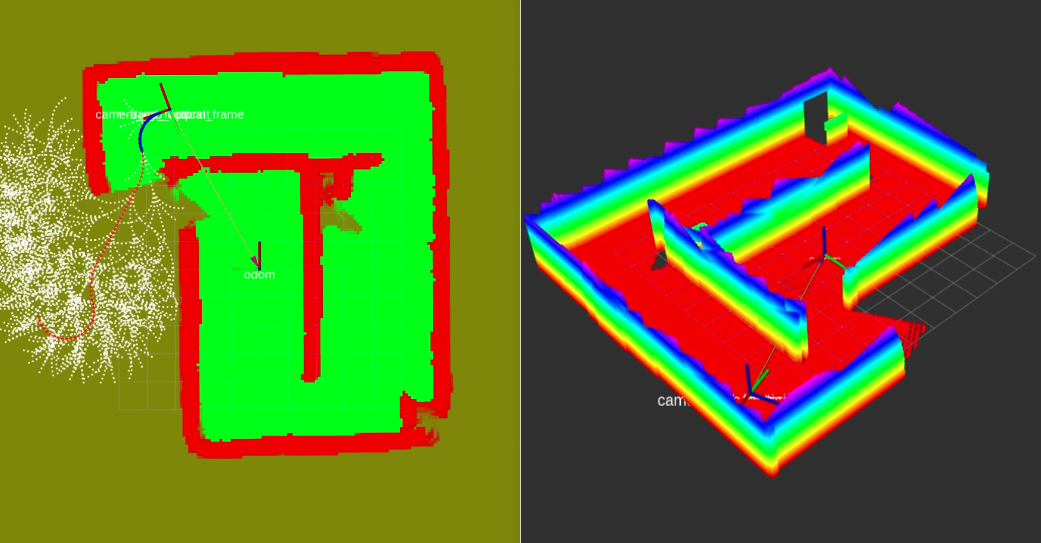}\\(g)
        \label{fig:test_7}
    \end{minipage}
    \begin{minipage}[tb]{0.5\linewidth}
        \centering
        \includegraphics[width=0.98\linewidth]{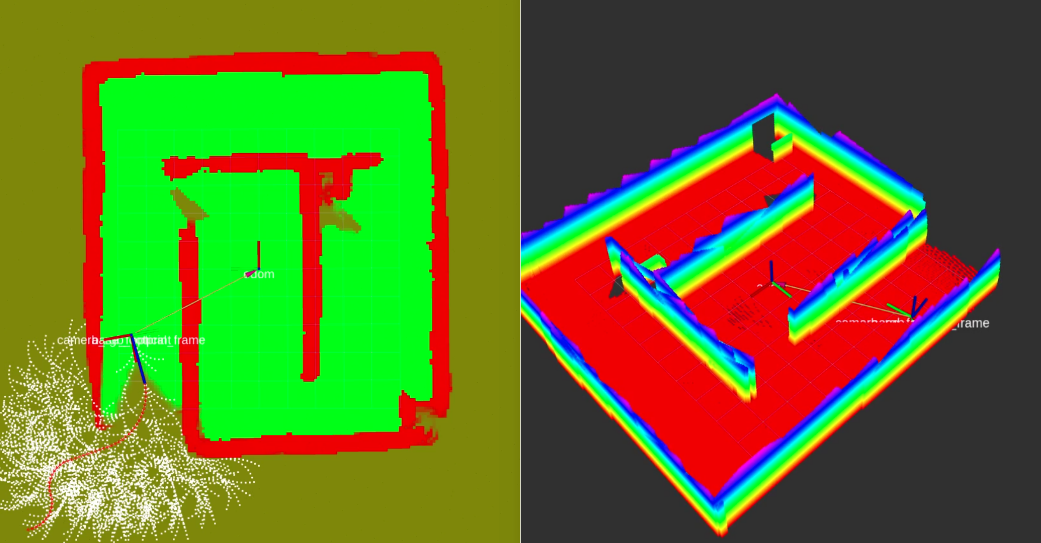}\\(h)
        \label{fig:test_8}
    \end{minipage}
    \caption{\small 
            3D Exploration process on top of Bayesian Hilbert Map with ParetoMCTS. \vspace{-10pt}
    }
    \label{fig:exploration}
\end{figure}

Our experiments are conducted in a Gazebo simulator and the environment to be mapped is shown in Fig.~\ref{fig:exploration_results}(a). The occupancy and 3D mapping results after exploration are shown in Fig.~\ref{fig:exploration_results}(b) and Fig.~\ref{fig:exploration_results}(c), respectively.
% In our experiment, for Bayesian Hilbert Mapping, we set hinge points resolution and query points resolution as 0.2 and 0.1, respectively. The kernel lengthscale $\gamma=40.0$. For ParetoMCTS, we set the maximum iteration for each search as 500.
Fig.~\ref{fig:exploration} shows the exploration process. % in our simulated environment. 
Specifically, the frontiers are detected by subtracting two consecutive occupancy maps (the bigger the discrepancy, the larger the reward of that frontier). Then we quantify the occupancy entropy as another rewarding metric. % for choosing which frontier the robot needs to reach. 
Typically, low-entropy values appear in explored areas while high-entropy values appear in unexplored areas. %In our experiment, all unexplored areas have the same entropy value. 
Therefore, subtrees with more branches stick out of the unknown area enjoy higher entropy reward.
% Since ParetoMCTS evaluate paths using cumulative reward, those path that have longer sections falling into unexplored areas will be assigned with higher accumulative rewards.
This means that the resulting path is likely to encourage the robot to uncover more unknown areas.
% With this observation, we could expect a path that has a higher accumulative reward will lead the robot to cover more unexplored areas.
This behavior is achieved in our experiment.
In Fig.~\ref{fig:exploration}(a), before planning, the robot could move either up~(and then left) or down since both directions have frontiers.
Our system chooses the latter direction~(see Fig.~\ref{fig:exploration}(b)), which turns out to be a right decision. Although the up direction has more and nearer frontiers, the unexplored areas beyond those frontiers are smaller than that of our selected frontiers.

\vspace{-15pt}
\section{Main Experimental Insights}\label{sec:insights}%\vspace{-10pt}
%=== Real time Bayesian Hilbert Map ===
The quality and computation time of BHM highly rely on the choice of kernel bandwidth and number of hinge points.
To make BHM more applicable for real-time scenarios, we propose a convex-hull-based hinge points selection strategy.
% Before the exploration task starts, a set of hinge points are defined in the whole area which is to be explored.
% For each time step, a convex hull is formed based on the training points.
At each time step, a slightly enlarged convex hull is built based on the training points within the field of view or sensing radius.
This tighter ``bounding box'' will enclose the most relevant hinge points.
We generate a more compact feature vector using these selected hinge points and only predict the occupancy probability locally.
% The hull-restricted hinge points will generate a new feature vector for predicting the local occupancy probability. 

%The hinge points which are not within the area covered by the training data will be discarded for the current local prediction. 
% By only including the important hinge points in the training and prediction process, the computation time is significantly decreased.
% These local prediction values will be fused into a global map in a way that no computation is needed for explored areas.

%=== ParetoMCTS ===
In the child node selection of ParetoMCTS, there is a weight parameter that balances exploitation of the recently discovered most promising child node and exploration of alternatives which may turn out to be a superior choice at later time.
Normalizing the exploitation scores or multiplying the exploration weight by the maximum exploitation score makes the weight parameter insensitive to the scales of objectives. 
% To take advantage of the continuous occupancy prediction property of BHM, we do not need to predict the occupancy values on a predefined grid and plan on this grid.
% Instead, we just need to predict on a specific position when the planner needs the occupancy information.
In addition, when the tree enters unknown areas, everywhere seems equally uncertain or informative, which makes the tree unable to grow asymmetrically, limiting the searching depth.
% We believe changing the tree growth mode from MCTS style to Rapidly-Exploring Random Tree (RRT) style will improve the estimation of the informativeness of unknown area, which is one of tasks that we plan to test and compare. 
% \input{5_experiments}
\newpage
\bibliographystyle{unsrt}
% \vspace{-15pt}
% {\small
\bibliography{references}

\begin{thebibliography}{10}

\bibitem{yamauchi1997frontier}
Brian Yamauchi.
\newblock A frontier-based approach for autonomous exploration.
\newblock In {\em Proceedings 1997 IEEE International Symposium on
  Computational Intelligence in Robotics and Automation CIRA'97.'Towards New
  Computational Principles for Robotics and Automation'}, pages 146--151. IEEE,
  1997.

\bibitem{burgard2000collaborative}
Wolfram Burgard, Mark Moors, Dieter Fox, Reid Simmons, and Sebastian Thrun.
\newblock Collaborative multi-robot exploration.
\newblock In {\em Proceedings 2000 ICRA. Millennium Conference. IEEE
  International Conference on Robotics and Automation. Symposia Proceedings
  (Cat. No. 00CH37065)}, volume~1, pages 476--481. IEEE, 2000.

\bibitem{shen2012stochastic}
Shaojie Shen, Nathan Michael, and Vijay Kumar.
\newblock Stochastic differential equation-based exploration algorithm for
  autonomous indoor 3d exploration with a micro-aerial vehicle.
\newblock {\em The International Journal of Robotics Research},
  31(12):1431--1444, 2012.

\bibitem{cieslewski2017rapid}
Titus Cieslewski, Elia Kaufmann, and Davide Scaramuzza.
\newblock Rapid exploration with multi-rotors: A frontier selection method for
  high speed flight.
\newblock In {\em 2017 IEEE/RSJ International Conference on Intelligent Robots
  and Systems (IROS)}, pages 2135--2142. IEEE, 2017.

\bibitem{charrow2015information}
Benjamin Charrow, Sikang Liu, Vijay Kumar, and Nathan Michael.
\newblock Information-theoretic mapping using cauchy-schwarz quadratic mutual
  information.
\newblock In {\em 2015 IEEE International Conference on Robotics and Automation
  (ICRA)}, pages 4791--4798. IEEE, 2015.

\bibitem{zhang2019fsmi}
Zhengdong Zhang, Trevor Henderson, Vivienne Sze, and Sertac Karaman.
\newblock Fsmi: Fast computation of shannon mutual information for
  information-theoretic mapping.
\newblock In {\em 2019 International Conference on Robotics and Automation
  (ICRA)}, pages 6912--6918. IEEE, 2019.

\bibitem{francis_functional_2020}
Gilad Francis, Lionel Ott, and Fabio Ramos.
\newblock Functional {Path} {Optimisation} for {Exploration} in {Continuous}
  {Occupancy} {Maps}.
\newblock In Nancy~M. Amato, Greg Hager, Shawna Thomas, and Miguel
  Torres-Torriti, editors, {\em Robotics {Research}}, Springer {Proceedings} in
  {Advanced} {Robotics}, pages 859--875, Cham, 2020. Springer International
  Publishing.

\bibitem{francis_occupancy_2019}
Gilad Francis, Lionel Ott, Roman Marchant, and Fabio Ramos.
\newblock Occupancy map building through {Bayesian} exploration:.
\newblock {\em The International Journal of Robotics Research}, May 2019.

\bibitem{bai_information-theoretic_2016}
Shi Bai, Jinkun Wang, Fanfei Chen, and Brendan Englot.
\newblock Information-theoretic exploration with {Bayesian} optimization.
\newblock In {\em 2016 {IEEE}/{RSJ} {International} {Conference} on
  {Intelligent} {Robots} and {Systems} ({IROS})}, pages 1816--1822, October
  2016.
\newblock ISSN: 2153-0866.

\bibitem{jadidi_mutual_2015}
Maani~Ghaffari Jadidi, Jaime~Valls Miro, and Gamini Dissanayake.
\newblock Mutual information-based exploration on continuous occupancy maps.
\newblock In {\em 2015 {IEEE}/{RSJ} {International} {Conference} on
  {Intelligent} {Robots} and {Systems} ({IROS})}, pages 6086--6092, September
  2015.

\bibitem{senanayake2017bayesian}
Ransalu Senanayake and Fabio Ramos.
\newblock Bayesian hilbert maps for dynamic continuous occupancy mapping.
\newblock In {\em Conference on Robot Learning}, pages 458--471, 2017.

\bibitem{ramos2016hilbert}
Fabio Ramos and Lionel Ott.
\newblock Hilbert maps: scalable continuous occupancy mapping with stochastic
  gradient descent.
\newblock {\em The International Journal of Robotics Research},
  35(14):1717--1730, 2016.

\bibitem{chen2019pareto}
Weizhe Chen and Lantao Liu.
\newblock Pareto monte carlo tree search for multi-objective informative
  planning.
\newblock In {\em Robotics: Science and Systems}, 2019.

\bibitem{hornung2013octomap}
Armin Hornung, Kai~M Wurm, Maren Bennewitz, Cyrill Stachniss, and Wolfram
  Burgard.
\newblock Octomap: An efficient probabilistic 3d mapping framework based on
  octrees.
\newblock {\em Autonomous robots}, 34(3):189--206, 2013.

\end{thebibliography}
% }
\end{document}